\documentclass[journal]{IEEEtran}

\usepackage[dvipsnames, svgnames, x11names]{xcolor}
\usepackage{xcolor,soul,framed} 

\colorlet{shadecolor}{yellow}
\usepackage[pdftex]{graphicx}
\graphicspath{{../pdf/}{../jpeg/}}
\DeclareGraphicsExtensions{.pdf,.jpeg,.png}

\usepackage[cmex10]{amsmath}

\usepackage{array}
\usepackage{mdwmath}
\usepackage{mdwtab}
\usepackage{eqparbox}
\usepackage{url}
\usepackage{booktabs}
\usepackage{amssymb}
\usepackage{multirow}

\begin{document}
\bstctlcite{IEEEexample:BSTcontrol}
    \title{Exploring Inter-frequency Guidance of Image for Lightweight Gaussian Denoising}
    
    \author{
    \IEEEauthorblockN{Zhuang Jia}
    \IEEEauthorblockA{ \\ Xiaomi Camera
    \\jiazhuang@xiaomi.com}
	}

\maketitle

\begin{abstract}
Image denoising is of vital importance in many imaging or computer vision related areas. With the convolutional neural networks showing strong capability in computer vision tasks, the performance of image denoising has also been brought up by CNN based methods.
Though CNN based image denoisers show promising results on this task, most of the current CNN based methods try to learn the mapping from noisy image to clean image directly, which lacks the explicit exploration of prior knowledge of images and noises. 
Natural images are observed to obey the reciprocal power law, implying the low-frequency band of image tend to occupy most of the energy. Thus in the condition of AGWN (additive gaussian white noise) deterioration, low-frequency band tend to preserve a higher PSNR than high-frequency band.
Considering the spatial morphological consistency of different frequency bands, low-frequency band with more fidelity can be used as a guidance to refine the more contaminated high-frequency bands.
Based on this thought, we proposed a novel network architecture denoted as IGNet, in order to refine the frequency bands from low to high in a progressive manner. Firstly, it decomposes the feature maps into high- and low-frequency subbands using DWT (discrete wavelet transform) iteratively, and then each low band features are used to refine the high band features. Finally, the refined feature maps are processed by a decoder to recover the clean result.
With this design, more inter-frequency prior and information are utilized, thus the model size can be lightened while still perserves competitive results. Experiments on several public datasets show that our model obtains competitive performance comparing with other state-of-the-art methods yet with a lightweight structure.

\end{abstract}


%
\IEEEpeerreviewmaketitle


\begin{figure} 
\centering 
\includegraphics[width=0.5\textwidth]{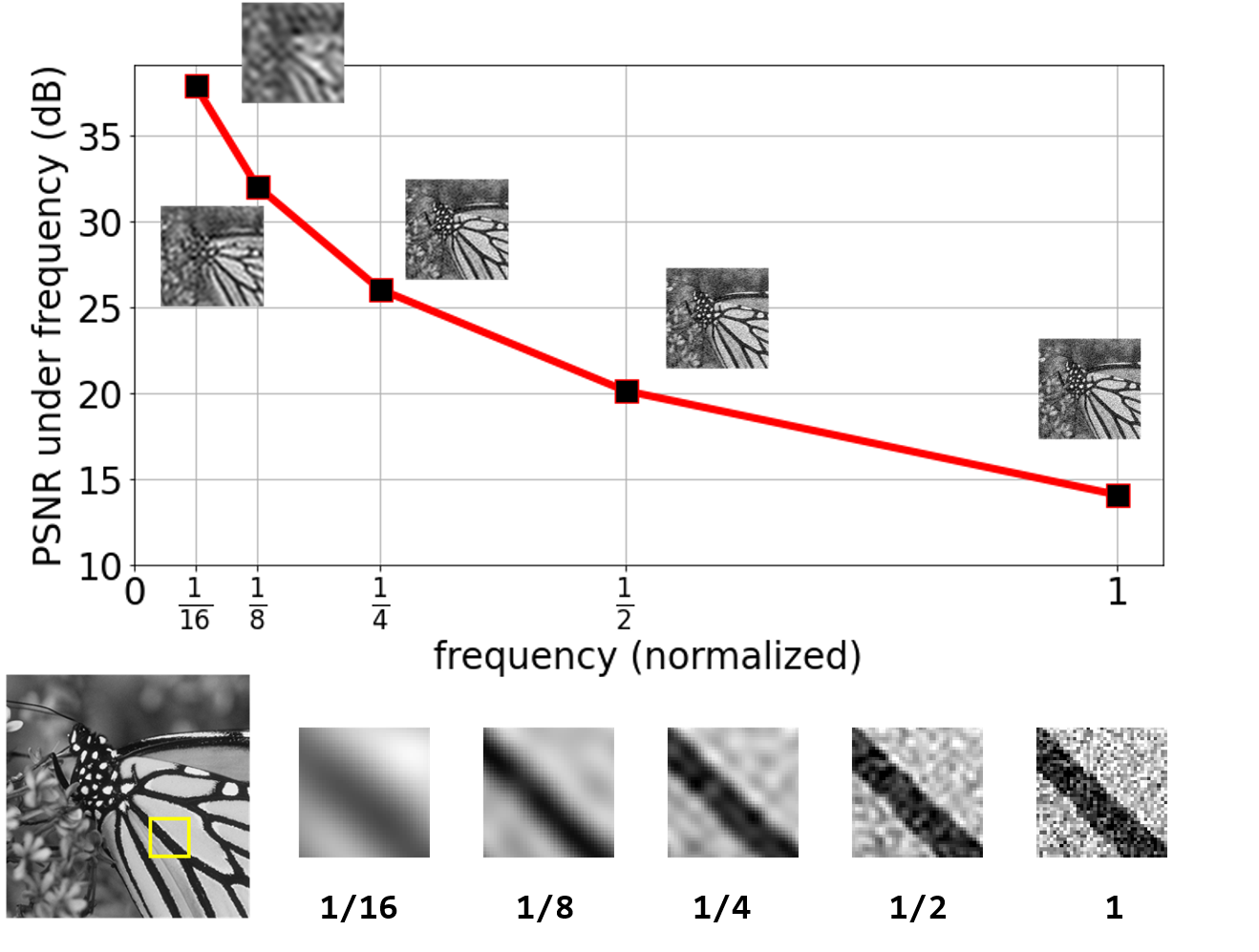} 
\caption{PSNR in different frequency limits of a noisy image ($\sigma=50$). The x-axis of above figure means the cutoff frequency in Fourier domain. Information under the cutoff frequency are preserved and over are discarded. This shows the PSNR increases with the decrease of frequency. Below shows a part of image limited in different frequencies, showing the inter-frequency morphological similarity between them which can be explored for guidance in denoising task.} 
\label{fig_psnr_vs_freq} 
\end{figure}

\section{Introduction}

\IEEEPARstart{I}{mage} denoising is a classic but still challenging task in image processing and computer vision field. Additive white Gaussian noise (AWGN) is a simple and effective modeling of random noise and has been widely studied in many researches. A lot of different methods and insights have been brought up to deal with this task. Denoising task can be regarded as an inverse problem which aims at recovering the clean image before unpredictable random noise was added using some prior knowledge, such as smoothness and edge preservation \cite{rudin1992nonlinear, osher2005iterative, he2010guided, tomasi1998bilateral}, self-similarity \cite{dabov2007image, buades2005non} , sparsity \cite{elad2006imagecvpr, mairal2007sparse}, and properties in transform domains \cite{starck2002curvelet, portilla2003image, mihcak1999low, chang2000adaptive}. Recently, with the rapid advancement of deep learning and convolutional neural networks (CNN), CNN-based methods have shown great improvement in many computer vision tasks including image classification, semantic segmentation, object detection, tracking \cite{szegedy2015going, he2016deep, chen2017rethinking, wang2020deep, he2017mask, redmon2018yolov3, bertinetto2016fully} and so forth. Besides the success of CNN-based methods for high-level image recognition tasks, image processing and low-level vision tasks, e.g. image denoising, super resolution, image inpainting also benefit from CNN models \cite{zhang2017beyond, mao2016image, zhang2018ffdnet, zhang2017learning, liu2018multi, dong2015image, shi2016real, ledig2017photo, yu2018generative}. In the view of low level tasks, CNN based models in fact act as a set of complicated learnable filters which can learn the mapping from deteriorated images (noisy, low resolution, scribbled etc.) to clean target. With the strong learning ability, these models easily surpass the delicately designed traditional methods when trained on a large properly collected and organized dataset.

Common CNN-based Gaussian denoisers use deep architecture which provides a more complex non-linearity and a larger receptive field, helping the model to learn how to remove the noises by understanding its semantic meanings within the receptive field. This architecture and learning scheme make it possible for models to capture the task-related image prior implicitly. In order to capture more useful information, a deeper and heavy network and more training data are required. However, this adaptive learning process may not be enough to explore the priors for Gaussian denoising task, and simply increasing the network capacity also brings more burden in computational cost, and may get to the saturation of performance. Therefore, exploring to explicitly apply image prior should be of consideration to boost the performance of CNN models and lighten the network parameters.

As for natural images, it has been observed and studied that their energy distribution along frequecy domain obeys the reciprocal power law, which means the energy of an natural image ususally tend to decrease dramatically when the frequency increases, causing the visual smoothness in the spatial domain \cite{tolhurst1992amplitude}. So when the noise shows the characteristics like AWGN, which is near to uniformly distributed in frequency domain, then it is obvious that the low-frequency part of an image tends to have a much higher signal-to-noise ratio (SNR) than its high-frequency conterpart, as shown in Fig.\ref{fig_psnr_vs_freq}. Moreover, under common circumstances, different frequency bands share the same or similar spatial morphological properties (edge direction, curvature etc.). This denoising task related prior knowledge should be taken into account for designing a more effective and lightweight CNN network.

In this paper, we proposed a novel network architecture based on the above stated observations and considerations. We explore the inter-frequency relations for the network learning, where the features from lower frequency band with more fidelity serve as guidance to recover the higher frequency band features. Firstly, we extract feature maps from input image using a simple feature extractor network which consists of stacked convolutional layers. Then, the feature maps are decomposed into different frequency bands using discrete wavelet transform (DWT), obtaining three high frequency subbands (i.e. $HL, LH, HH$) and one low frequency subband (i.e. $LL$). The low-frequency subband features are then used to refine the high-frequency subbands on the one hand, and on the other hand decomposed once more using DWT into another stage, like in wavelet pyramid method. These steps are taken progressively, making each high-frequency components refined by all the lower frequency subbands. After all subbands have been refined, inverse discrete wavelet transform (IDWT) are conducted to merge them into full-frequency band features, and then decoder network which follows the refinement procedure performs recovery of the clean image from the refined feature maps. 

The main contributions of this paper can be summarized in 3-fold as follows:

\begin{itemize}

\item The feasibility of inter-frequency guidance of image for Gaussian denoising has been analysed and experimented, showing that the split and separately processing of high- and low-frequency bands, as well as refinement of higher frequency bands using lower frequency information can be beneficial to construct a more compact network.

\item A novel network architecture is presented to refine the more contaminated higher frequency components progressively using lower frequency components with more fidelity by exploring the inter-frequency band relations. 

\item Due to the advantage of constraint from inter-frequency prior, the proposed model can be reduced to a relatively small size, yet still show promising results in experiments with different test sets and different noise levels. 

\end{itemize}

The remainder of paper is organized as follows. Sec II briefly reviews the traditional and CNN-based Gaussian denoising methods. Sec III introduces the proposed method and network architecture. Sec IV provides the experiment results for verifying the effectiveness and evaluating the performance of proposed method. Sec V concludes the paper.

\section{Related Work}

In this section, a concise review of Gaussian noise removal using both traditional and learning based methods are presented to illustrate the development of this task. In addition, the frequency information and its applications in CNN-based methods are also briefly surveyed, which shows the effectiveness and improvement using the frequency domain prior knowledges in image processing and computer vision tasks.

\subsection{Gaussian Denoising}
As for the effectiveness and simplicity for analysis, Gaussian denoising of images has been studied for a long time. Traditional methods tackle this problem in several technical ways, such as self-similarity, sparsity representation, domain transform etc. . For example, the classic bilateral filter method takes both spatial position similarity and value similarity into account, which makes the results to be more edge-preserving while supress the noise contamination \cite{tomasi1998bilateral}. Dictionary learning and sparse representation has also been explored in denoising task as in \cite{elad2006image}, which uses K-SVD to learn a effective dictionary of image contents. Non-local means \cite{buades2005non} and BM3D \cite{dabov2007image} are the representatives of non-local self similarity based methods, which reduce the noise level by averaging similar patches from different regions. BM3D surpasses many traditional methods in many practical occasions and has set a strong baseline for image denoising. Moreover, TNRD \cite{chen2016trainable} proposed a flexible learning framework via nonlinear reaction diffusion models.

Recent years have witness the burst of deep learning and convolutional neural networks, which has deeply changed the paradigm of computer vision and image processing area. Many CNN-based approaches for image denoising have been proposed. DnCNN \cite{zhang2017beyond} experimentally analysed the effect of batch normalization (BN) and residual learning scheme in denoising task, and designed a new network architecture to obtain a competitive result comparing with traditional state-of-the-art methods. Residual Encoder-Decoder Network(RED) uses encoder-decoder architecture, and introduced symmetric skip connections to compensate the loss of fine-grain details in encoder-decoder process, which is of vital importance for low-level vision tasks \cite{mao2016image}. Inspired by the successes of CNN-based methods in Gaussian denoising task, several novel CNN-based models for denoising and other image restoration tasks are proposed \cite{zhang2018ffdnet, liu2018multi, zhang2020rdnir}. With the development of Transformer-based networks, the capability of Transformer-based models on low-level vision tasks have also been studied and experimented \cite{chen2021pre, liang2021swinir}.

\subsection{Frequency Prior in CNN-based Models}

Despite the progress made by CNN-based methods which mainly rely on the strong capability of CNN network to learn the prior knowledge implicitly using a large training dataset, there are also researches to explicitly analyse and utilize the prior knowledge about the frequency characteristics of natural images. For example, DualCNN proposes a two branch network structure to reconstruct the structure and fine-details separately, and combine them to form a more edge-preserving output \cite{pan2018learning}, which takes into account the difference between relatively low-frequency structures and high-frequency details in network training. Multi-Wavelet CNN (MWCNN) uses discrete wavelet transform (DWT) and its inverse form (IDWT) to replace commonly used pooling and upscaling layers, which gives promising results in low-level tasks \cite{liu2018multi}. In addition to low-level task, frequency-related operations may also benefit recognition tasks. In \cite{zhang2018hartley}, a novel pooling layer using the low-pass filter in frequency domain via Hartley transform has been proposed, experiments show that it can preserve more features in downscale step than the common max pooling or average pooling, which can improve the performance in several recognition tasks.

These above mentioned works have proven the effectiveness of frequency priors in image tasks. Our proposed method also consider to use the frequency prior to improve the model performance. Different from most of the previous works which refine different frequency components independently and combine them in serial or parallel manner, our idea is based on the inter-frequency similarity prior, where high-SNR low-frequency components serve as guidance to refine low-SNR high-frequency components. Therefore, we exploit the similarity and relations explicitly by fusing features from low- and high-frequency bands to refine the high-frequency features, forcing the communication within different frequency bands. As a result of more prior information, the number of learnable parameters can be significantly reduced while keeping a competitive performance.

\section{Method}

In this section, we first introduce the whole architecture of proposed method, which will be denoted as IGNet (\textbf{I}nter-frequency \textbf{G}uidance \textbf{Net}work) in the following statements for simplicity. Then the core module of IGNet, namely FPFR (Frequency Progressive Feature Refiner) is presented and described formally. The LGR (Low-frequency Guided Refiner) module which is used to combine the low- and high-frequency information for refining high-frequency subbands are described at last of this section.

\subsection{Network Architecture}

\begin{figure} 
\centering 
\includegraphics[width=0.5\textwidth]{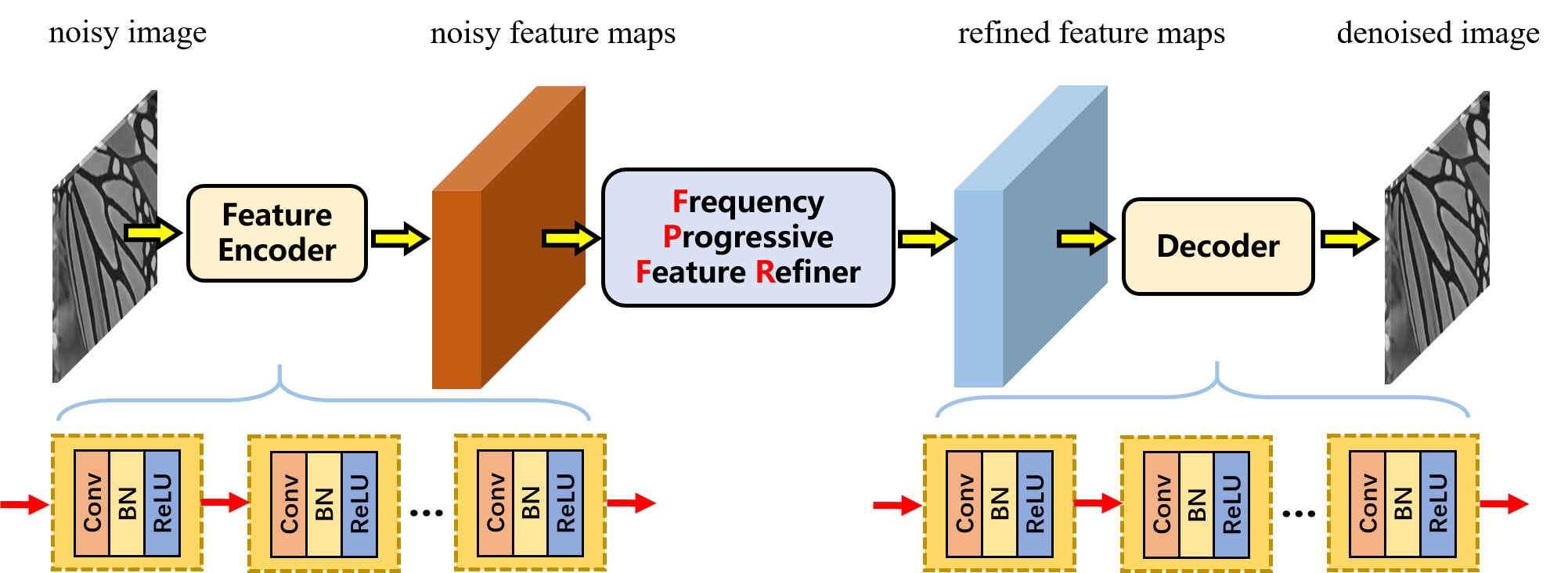} 
\caption{Architecture of proposed IGNet. FPFR module is inserted into feature extractor and decoder to modify the noisy feature maps using inter-frequency guidance.} 
\label{fig_ignet} 
\end{figure}

IGNet is mainly composed of three components. First of all, a feature extractor module is used to extract the feature maps of noisy input image, this module is implemented using a stack of [Conv+BN+ReLU] blocks, as in DnCNN model \cite{zhang2017beyond}. After the feature maps are extracted, a Frequency Progressive Feature Refiner (FPFR) module is followed to refine the feature maps using inter-frequency guidance by decomposing the feature maps into different frequency subbands, and the refinement process is conducted from low-frequency band to high-frequency band progressively. Then a decoder module with the same block formation as the feature extractor is used to decode the refined feature maps into a clean prediction image. The network architecture of IGNet is shown in Fig. \ref{fig_ignet}.

\subsection{FPFR Module}

FPFR module serves as the core component to utilize inter-frequency relations for the modification of noisy feature maps in a frequency-progressive way. Firstly, the extracted feature maps are decomposed by a series of DWT operations, then the low-frequency subband is fused to the high-frequency subbands correspondingly as a guidance to refine the noisy high-frequency feature maps. Then, the refined high-frequency features are combined with corresponding low-frequency features using IDWT, and used to guide their corresponding higher frequency counterparts. This procedure takes place for several times on features of different frequency subbands. Finally, after all the refinement and combination, FPFR outputs the refined full-frequency features.

In order to illustrate the operations in the FPFR module clearly and formally, we here describe the FPFR module operations in mathematical formation. Denote the input noisy image as $x$, and the feature extractor as FE, then 
\begin{align} 
f_x &= {\rm FE}(x)
\end{align}
then the first stage of DWT is conducted, decomposing the feature maps $f_x$ into one low-frequency part and three high-frequency part ( high frequency in height direction, width direction and both):
\begin{align}
LL_0, \{ HL_0, LH_0, HH_0\} &= {\rm DWT}(f_x)
\end{align}
for simplicity of notification, we use $L$ to denote $LL$ and $H$ to denote $\{ HL, LH, HH\}$. After the subbands $L_0$ and $H_0$ are separated, the first refinement starts by taking $L_0$ to guide $H_0$:
\begin{align}
H_0^{(0)} &= {\rm LGR}(L_0, H_0)
\end{align}
where LGR refers to Low-frequency Guided Refiner, which fuse corresponding low- and high-frequency subbands and output refined high-frequency subbands. The superscript $(0)$ in $H_0^{(0)}$ means it is the 0-th (first) refinement for original noisy $H_0$.

Then the lower subband $L_0$ is decomposed using DWT as before, and resulting into the second stage of DWT.
\begin{align}
 L_1, H_1 &= {\rm DWT}(L_0) 
\end{align}
Like the first stage, $L_1$ is used to refine $H_1$ via LGR module.
\begin{align}
H_1^{(0)} &= {\rm LGR}(L_1, H_1)
\end{align}
the refined $H_1^{(0)}$ has the same size as $H_1$, so it can be regarded as a cleaner version of high-frequency features, then we use $H_1^{(0)}$ and $L_1$ to reconstruct by inverse discrete wavelet transform (IDWT) to form a modified $L_0$ where the two components $H_1$ and $L_1$ come from, and use this modified $L_0$ features to guide the refinement $H_0$, that is:
\begin{align}
L_0^{(0)} &= {\rm IDWT}(L_1, H_1^{(0)}) \\
H_0^{(1)} &= {\rm LGR}(L_0^{(0)}, H_0) \\
\end{align}
Similarly, we can get to the third stage, and progressively refine the high-frequency features from bottom to top. These steps can be presented as the following formulas:
\begin{align}
L_2, H_2 &= {\rm DWT}(L_1) \\
H_2^{(0)} &= {\rm LGR}(L_2, H_2) \\
L_1^{(0)} &= {\rm IDWT}(L_2,  H_2^{(0)}) \\
H_1^{(1)} &= {\rm LGR}(L_1^{(0)}, H_1^{(0)}) \\
L_0^{(1)} &= {\rm IDWT}(L_1^{(0)},  H_1^{(1)}) \\
H_0^{(2)} &= {\rm LGR}(L_0^{(1)}, H_0^{(1)})
\end{align}
At last, the lowest frequency subband $L_2$ is refined by a convolution layer.
\begin{align}
L_2^{recon} &= {\rm LT}(L_2)
\end{align}

After all the frequency bands are refined as above, ${\rm IDWT}$ is progressively applied to reconstruct the full-frequency band feature maps.
\begin{align}
L_1^{recon} &= {\rm IDWT}(L_2^{recon}, H_2^{(0)}) \\
L_0^{recon} &= {\rm IDWT}(L_1^{recon}, H_1^{(1)}) \\
f_x^{recon} &= {\rm IDWT}(L_0^{recon}, H_0^{(2)}) 
\end{align}

\begin{figure} 
\centering 
\includegraphics[width=0.5\textwidth]{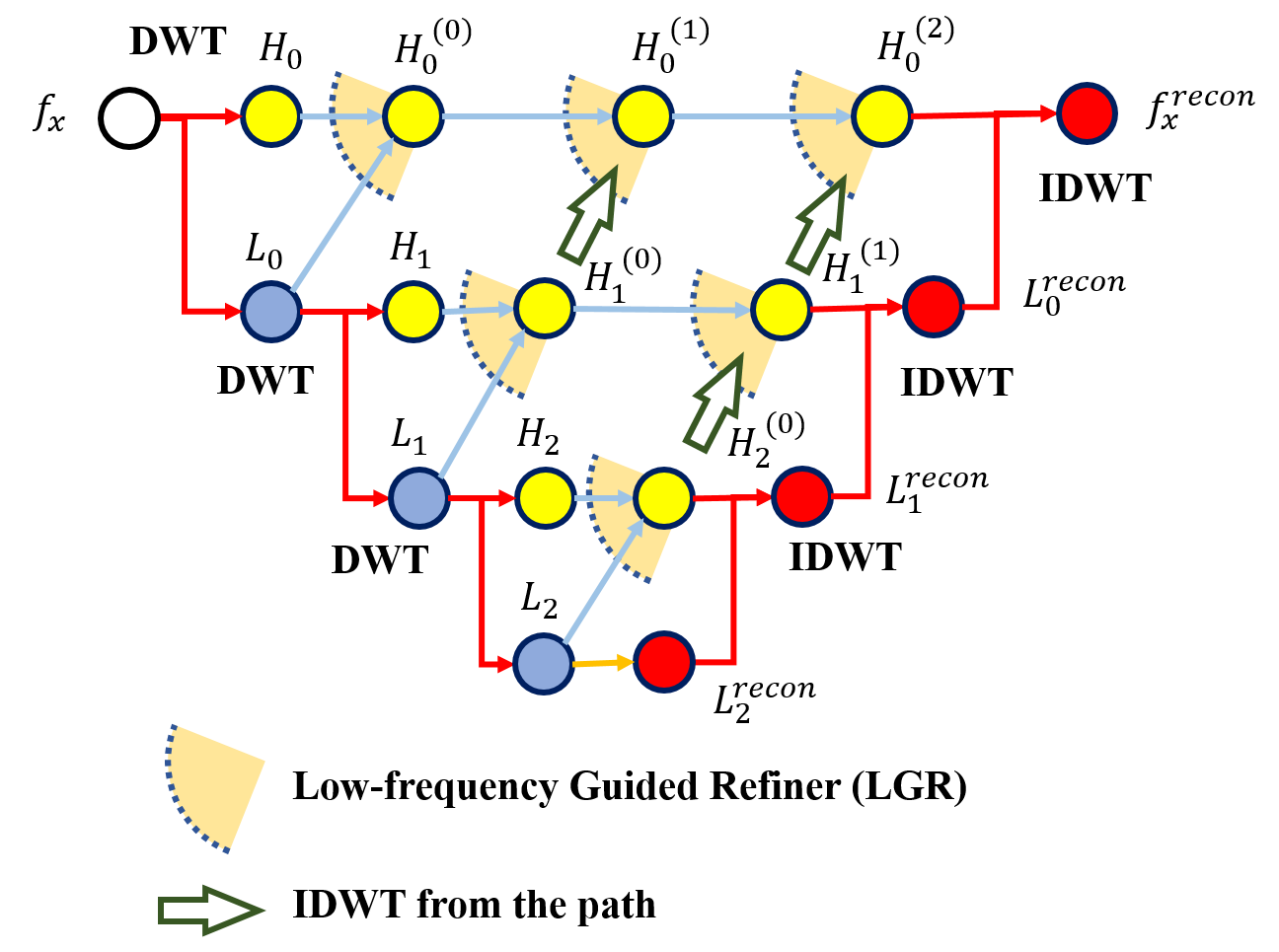} 
\caption{Implementation of FPFR module. Low frequency feature guidance is applied to refine features with higher frequency via LGR module. The refined features of each frequency band are finally combined using } 
\label{fig_fpfr} 
\end{figure}

\begin{figure} 
\centering 
\includegraphics[width=0.5\textwidth]{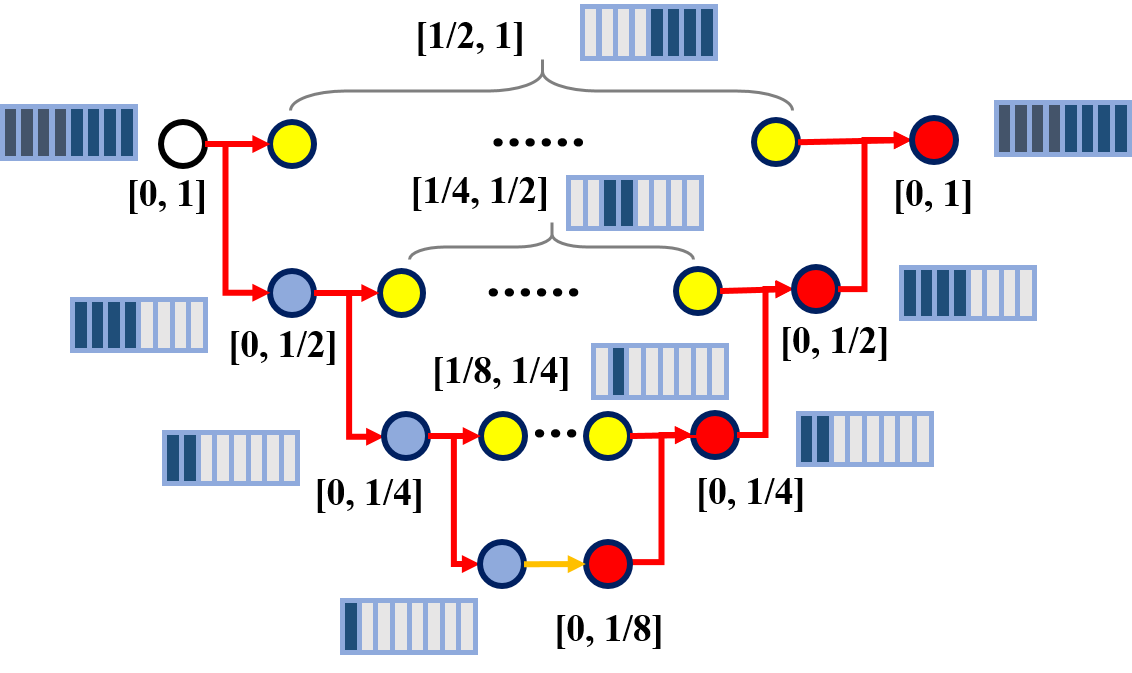} 
\caption{Frequency band of each feature map in FPFR module. For example, the first LL subband is denoted as $[0, 1/2]$, and subbands \{$HL$, $LH$, $HH$\} are all denoted as [1/2, 1] which shows the frequency band of each subband in width, height and both dimensions.} 
\label{fig_fpfr_freq} 
\end{figure}

To be clear about the frequency bandwidth in each stage, Fig. \ref{fig_fpfr_freq} is presented, where we denote the full-frequency band as $[0, 1]$, and the frequency bands of $LL$ and \{$HL$, $LH$, $HH$\} from it using DWT are $[0, 1/2]$ and $[1/2, 1]$ respectively. From this figure, we can see that the guidance of frequency band is as the following:
\begin{align}
[0, 1/2^{i}] \xrightarrow{}{} [1/2^{i}, 1/2^{i-1}]
\end{align}

\subsection{Low-frequency Guided Refiner}

\begin{figure} 
\centering 
\includegraphics[width=0.5\textwidth]{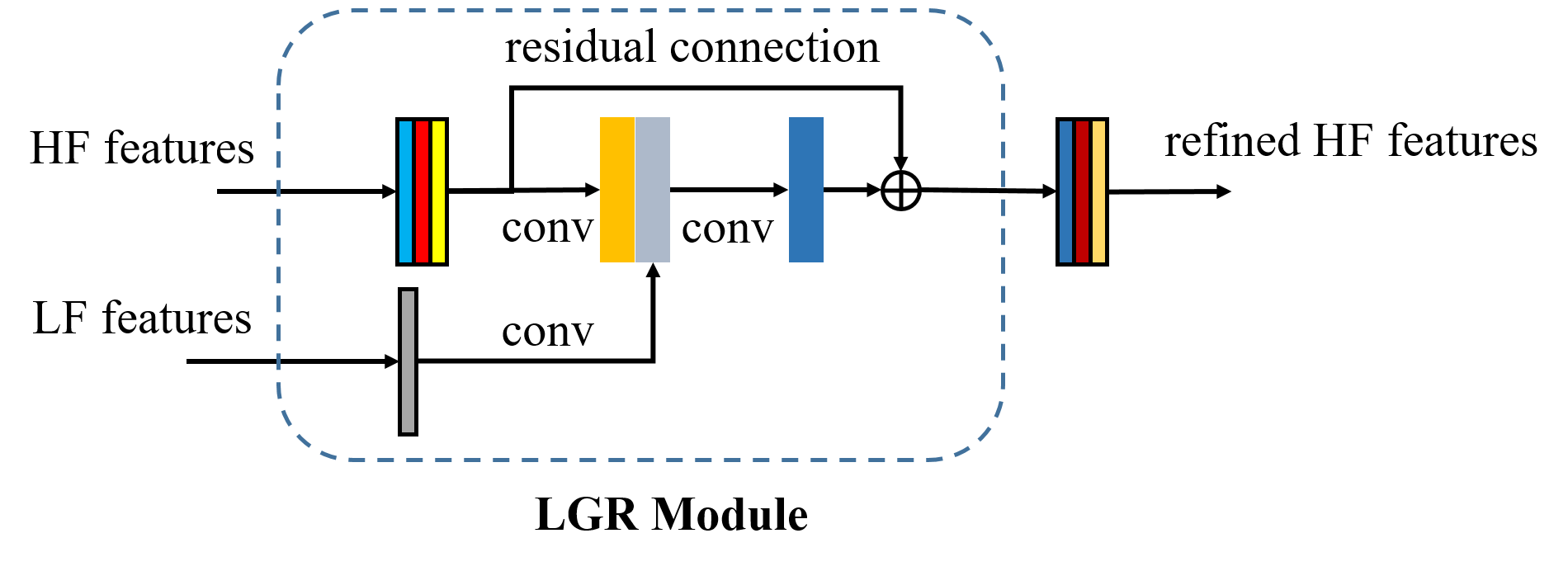} 
\caption{Structure of Low-frequency Guided Refiner (LGR) module. The low-frequency subband features are fused with the high-frequency subband features as a guidance to refine the high-frequency features. Residual structure is used here to make the output be more like original high-frequency features.} 
\label{fig_lgr} 
\end{figure}

In the above module, the fusion of low and high frequency features for high frequency feature refinement is based on LGR (Low-frequency Guided Refiner) Module. The structure of the LGR module is shown as Fig. \ref{fig_lgr}

The LGR takes two feature maps with same resolution and $3C$ and $C$ channels for high-frequency feature maps $HF$ and low-frequency $LF$ feature maps respectively. Both $HF$ and $LF$ are fed into a convolutional layer, and the outputs are fused together by concatenation along channel dimension. Then the fused features are processed by another convolutional layer to construct the inter-frequency relations between $HF$ and $LF$ features, and the output feature maps of this step are of $3C$ channels, which is the same as input $HF$ features. Then the results are added with original input $HF$ in a residual manner, because the target of this module is to refine the $HF$ features, this residual connection helps training when we expect the output $HF^{refined}$ to be more like the input $HF$. A more formal description is as follows:

\begin{align}
 f_{HF} &= {\rm Conv}_1(HF)  \\
 f_{LF} &= {\rm Conv}_2(LF)  \\
 res &= {\rm Conv}_3({\rm Concat}(f_{HF}, f_{LF}))  \\
 HF^{refined} &= HF + res
\end{align}

\section{Experiments}

In this section, the experiment results and analysis are reported. We firstly describe the basic settings of experiments. Then the analysis and visualization of each implementation are provided. Finally, we compare the proposed methods with current state-of-the-art CNN based methods, showing the effectiveness and efficiency in model size of proposed method.

\subsection{Experiment Settings}

Most of the previous works on denoising are tested on gray images, so we choose to train our models for grayscale image denoising, and evaluate the performance with public dataset for convinience of comparison. For training our model, we use DIV2K \cite{agustsson2017ntire}, Waterloo Exploration Database (WED) \cite{ma2016waterloo}, and Berkeley Segmentation Dataset (BSD) \cite{martin2001database} as train dataset as suggested in \cite{liu2018multi} to construct the training dataset. To be more specific, 800 images from DIV2K, 4744 images from WED and 200 images from BSD are combined together for our training. 

To evaluate our trained models, several widely used public datasets are employed as evaluation dataset. For testing our grayscale denoisers, Set12 \cite{zhang2017beyond}, BSD68 \cite{martin2001database} and Urban100 \cite{huang2015single} are adopted. 

In order to train the model, the images from combined training dataset are cropped as patches with size 128 $\times$ 128. For simplicity of comparison, the noise levels considered in our experiments are $\sigma = 15,\ 25$ and $50$. The Adam optimizer is used in our experiments with default settings and cosine annealing learning rate scheduler is used with initialized $lr = 1e-3$. For each noise level, the maximum training epoch is 120. Augmentations of rescale, flip, rotation are also utilized. The implementations are based on PyTorch, and all the experiments are conducted in Ubuntu 18.04 server with Nvidia A100 GPU.

\begin{figure*} 
\centering 
\includegraphics[width=0.9\textwidth]{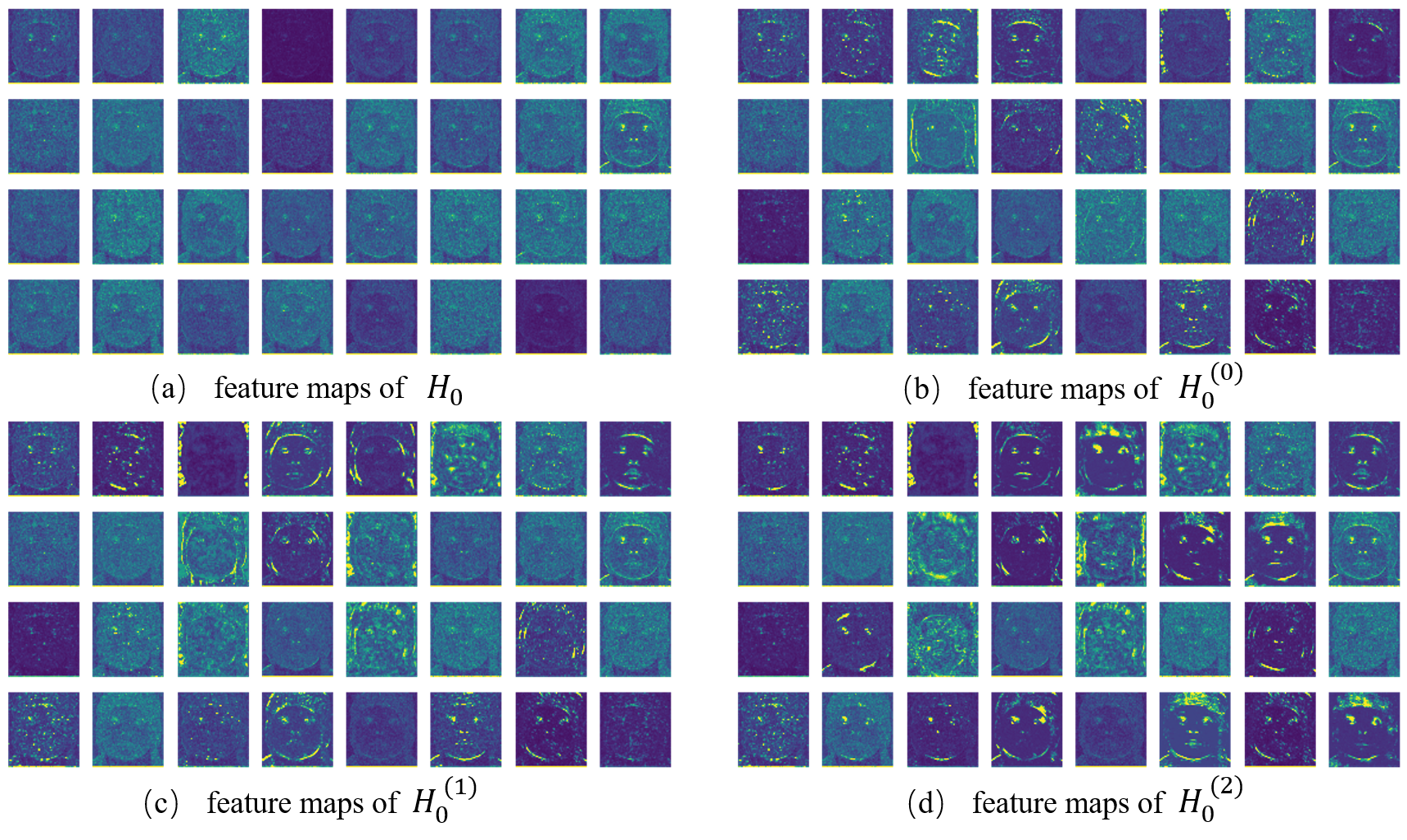} 
\caption{Feature map visualizations of the first stage of high frequency band $H_0$ and its refinements. (a) is the feature maps of $H_0$, (b) to (d) are $H_0^{(0)}$ to $H_0^{(2)}$. It can be seen from the comparison of each refinement result that low-frequency guidance is effective in modifying noisy high-frequency feature maps.} 
\label{fig_feat_refine} 
\end{figure*}

\subsection{Guided Refinement Visualization}

The core strength of our IGNet comes from the separately processing of low-frequency subband $LL$ and high-frequency subbands ${HL, LH, HH}$, and the inserted FPFR module where feature maps of high-frequency components are refined progressively. Therefore, we firstly analyse the refinement procedure of FPFR module via visualization of feature maps of each refinement step. As shown in Fig. \ref{fig_feat_refine}, the input is ``baby.png'' from Set5 dataset with $\sigma=50$ noise. The refinement results of the first DWT stage is selected because of more refinement steps. As the number of feature map channels from feature extractor is set to 32, so the original high-frequency subbands $H_0$ have total 96 channels. For convinience of display, we only show the middle 32 feature maps (i.e. the $LH$ part) for visualization and analysis, noticing that $HL$ and $HH$ feature maps are similar to the displayed part. 

As can be shown in Fig. \ref{fig_feat_refine}, the four subfigures are feature maps of original $H_0$, and its three consecutive refined results $H_0^{(0)}$, $H_0^{(1)}$ and $H_0^{(2)}$. All feature map channels are normalized and taken absolute value to better compare the differences.

In the $H_0$ feature maps, the noise with high frequency are obvious shown, and the effective image features are mixed together with noise which can hardly be discriminated. Then, in $H_0^{(0)}$ where $H_0$ is enhanced via guidance from $L_0$, its low-frequency counterpart, some of the feature maps begin to show more clear high-frequency features of image, yet more feature maps are still left noisy and unclear. In $H_0^{(1)}$, more feature maps have enhanced high-frequency edges, and the contrast in these feature maps increased, which is beneficial to reconstruct clean images. At last, after the refinement using guidance from all three stages, the feature maps of $H_0^{(2)}$ show more enhancement and sparsity. This analysis proves that the low-frequency guided refinement in FPFR module is effective in optimizing the high-frequency feature maps.

\subsection{Ablation Study}

In this subsection, to validate the effectiveness of the proposed method more concretely, ablation studies are conducted to analyse the influence of each structure. Firstly, we reduce the number of DWT stages, and keep other settings unchanged. The results are shown in Table. \ref{tab_ablation_stage}. $IGNet_1$ refers to the models with one DWT and IDWT, thus only first stage features ($H_0$) have been refined. $IGNet_{12}$  refers to the model structure with first and second stages ($H_0$ and $H_1$) used for refinement. IGNet is the proposed network architecture with all three stages for comparison. The test set used here is BSD68, and the noise level in the experiments are all set to $\sigma=50$. From Table. \ref{tab_ablation_stage} it can be seen that more stages of DWT, i.e. more times of frequency split, is beneficial to the denoising task. This result is mainly due to that more stages of DWT can explore more inter-frequency information and the lowest frequency band has higher SNR which can serve as a more robust guidance for the recovery of high-frequency components.

\begin{table}[]
\centering
\caption{Ablation study for number of DWT stages}
\label{tab_ablation_stage}
\begin{tabular}{lllll}
\toprule
          & $H_0$ & $H_1$ & $H_2$ & PSNR (dB) \\ \midrule
$IGNet_1$  & $\checkmark$  &      &      &      26.43     \\
$IGNet_{12}$ & $\checkmark$  & $\checkmark$  &      &     26.52      \\
$IGNet$     & $\checkmark$  & $\checkmark$  & $\checkmark$    &      \textbf{26.61}     \\ \bottomrule
\end{tabular}
\end{table}

\begin{figure} 
\centering 
\includegraphics[width=0.5\textwidth]{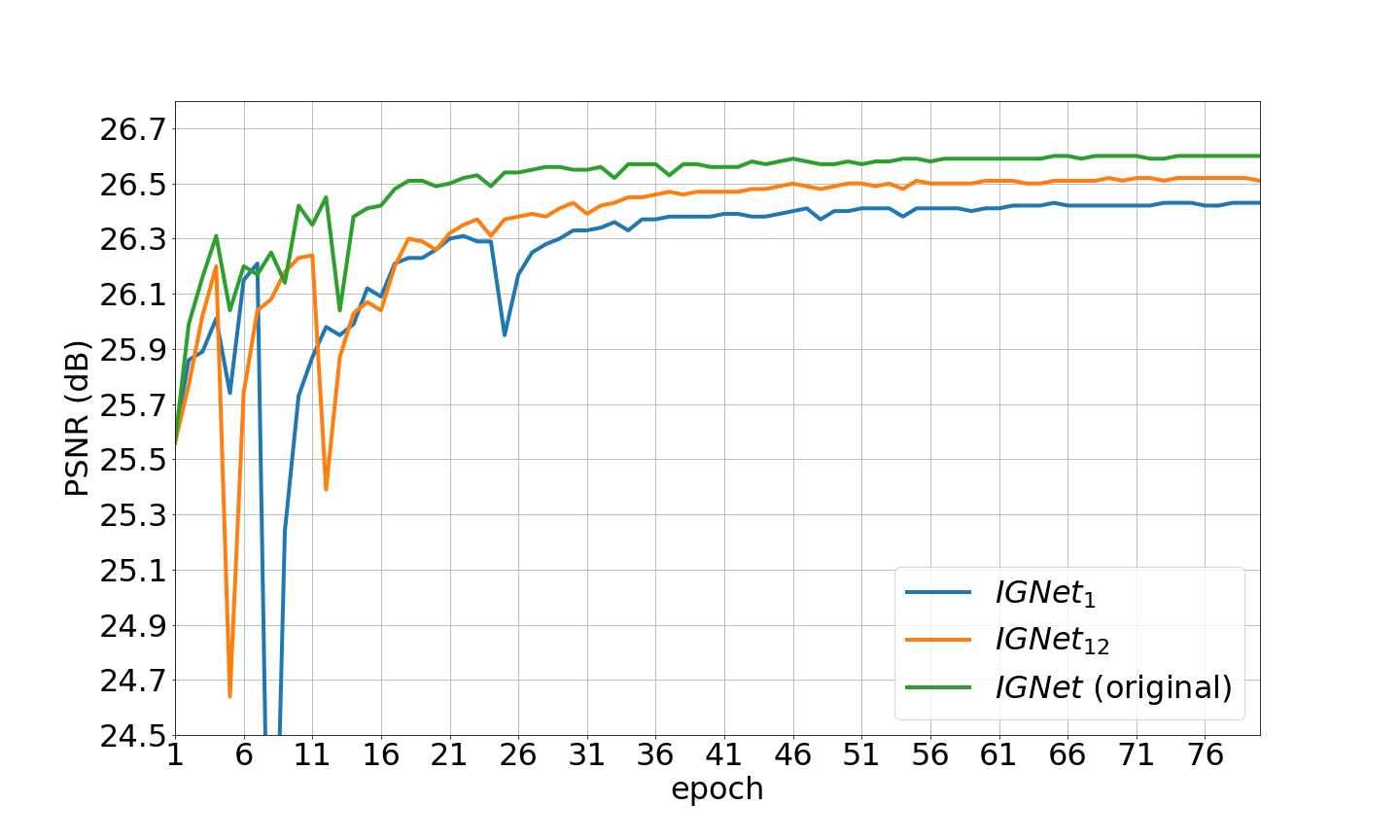} 
\caption{PSNR in each epoch with different number of DWT stages. $IGNet_1$ and $IGNet_{12}$ has 1 and 2 stages correspondingly, comparing $IGNet$ which has 3 DWT stages.} 
\label{fig_ablation_stage} 
\end{figure}

\begin{table}[]
\centering
\caption{Ablation study of inter-frequency guidance module and $LL$,\{$HL$, $LH$, $HH$ \} subbands separation}
\label{tab_ablation_lgr}
\begin{tabular}{@{}lllll@{}}
\toprule
               & Set5        & BSD68 & Urban100 \\ \midrule
baseline  & 27.52 / 0.7968 &   26.61 / 0.7371  &   27.12 / 0.8239     \\
w/o LGR module &   27.41 / 0.7930  &  26.53 / 0.7327  &  26.86 / 0.8146      \\ 
w/o L\&H band sep &   27.07 / 0.7804  &  26.36 / 0.7240  &  26.11 / 0.7904      \\ \bottomrule
\end{tabular}
\end{table}

\begin{figure} 
\centering 
\includegraphics[width=0.5\textwidth]{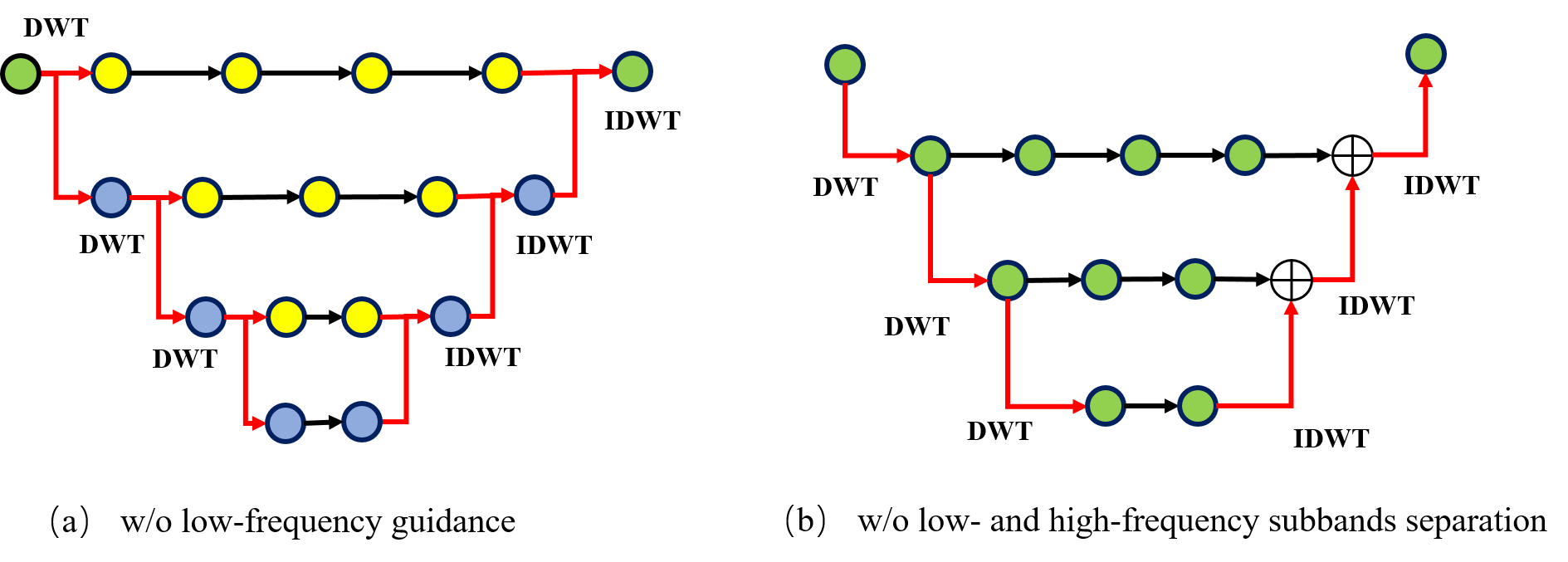} 
\caption{Alternative structures experimented as ablation study. (a) refers to without low-frequency guidance structure, and (b) refers to combining low- and high- frequency subbands together for next DWT instead of conducting DWT only for low-frequency subband in a pyramidic way.} 
\label{fig_ablation_lgr} 
\end{figure}

In addition, we also studied the effectiveness of LGR module, which combines the low- and high-frequency information for the refinement of high-frequency band features, and the separation of low-frequency subband $LL$ and high-frequency subbands ${HL, LH, HH}$. For the first comparison, we replace the LGR module with normal residual convolutional blocks which only operate on high-frequency feature maps, leaving the inter-frequency communication abandoned. For the second comparison, we concatenate $LL$ and ${HL, LH, HH}$ features together in each stage, and directly refine features from each DWT stage with convolutional layers, and then use IDWT to upscale features from each stage, and merge the refined features and upscaled features below by adding them together. The two alternative structures are shown in Fig. \ref{fig_ablation_lgr} to make it clearer. The results of the experiments are shown in Table. \ref{tab_ablation_lgr}. 

The row labeled with ``baseline'' refers to the results from proposed IGNet with both low- and high-band separation and LGR module, the row with ``w/o LGR module '' and `` w/o L\&H band sep '' refer to the results without LGR module and without low- and high-band processed separately as elaborated above ((a) and (b) in Fig. \ref{fig_ablation_lgr} respectively). From the results in Table. \ref{tab_ablation_lgr}, we can see that high- and low-frequency separation performs better results compared with simply merging all subbands together for the next DWT. In fact, using only low-frequency band for the next DWT not only boost the performance, but also reduces the number of parameters by reducing the number of feature map channels. Moreover, LGR module brings about 0.15dB in average, which implys the low-frequency information is a useful guidance for the the high-frequency feature modification, and splitting high- and low-frequency bands can also brings about 0.42 dB compared with merging all subbands together. It can be concluded from the experiment results that the architecture of IGNet is effective for learning a denoising task.

\subsection{Qualitative and Quantitative Comparisons}

\begin{figure} 
\centering 
\includegraphics[width=0.5\textwidth]{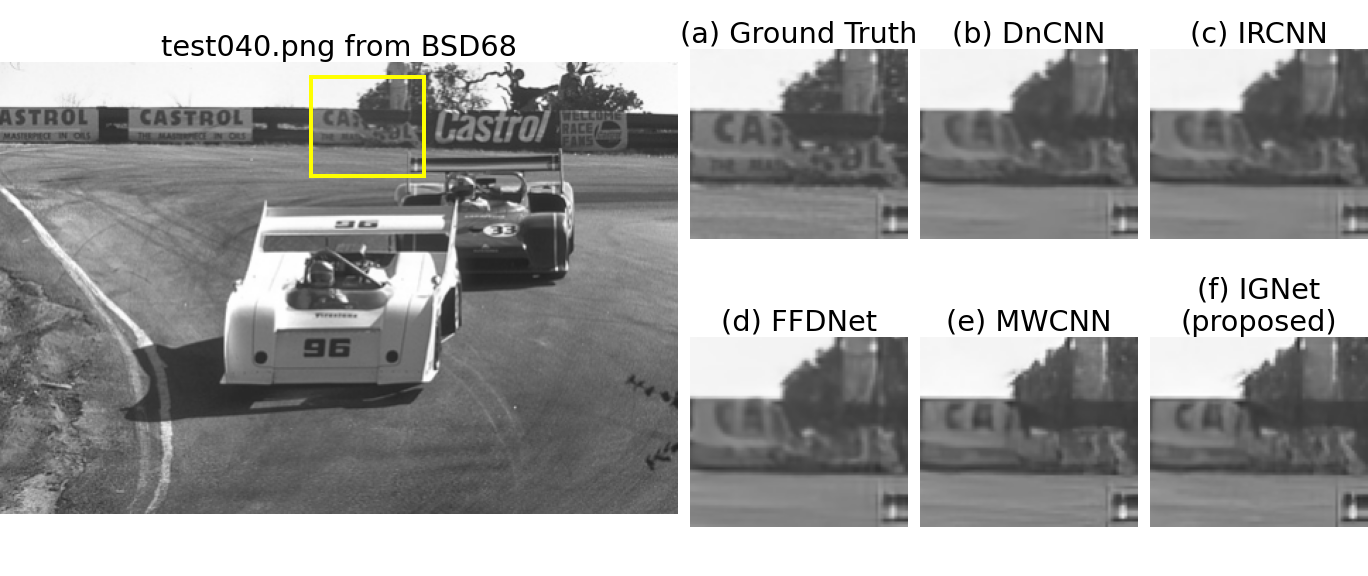} 
\caption{Comparison of results using different models on test\_040.png.} 
\label{fig_vis_1} 
\end{figure}

\begin{figure} 
\centering 
\includegraphics[width=0.5\textwidth]{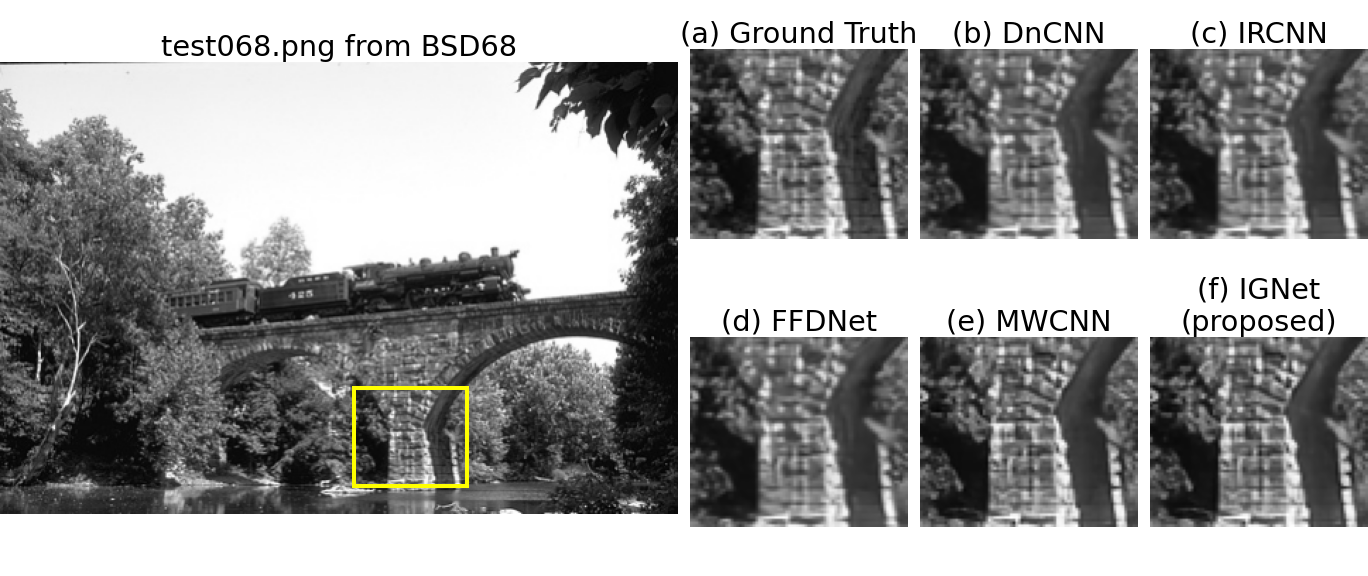} 
\caption{Comparison of results using different models on test\_068.png.} 
\label{fig_vis_2} 
\end{figure}

\begin{table*}[]
\scriptsize
\centering
\caption{Comparisons with current competing methods both traditional and CNN-based, metrics in PSNR(dB) and SSIM.}
\label{tab_sota_compare}
\begin{tabular}{@{}cccccccccc@{}}
\toprule
                          & sigma & BM3D           & TNRD           & DnCNN        & IRCNN    & FFDNet   &  MWCNN       & IGNet    & IGNet+      \\ \midrule
\multirow{3}{*}{Set12}    & 15    & 32.37 / 0.8952   & 32.50 / 0.8962   & 32.86 / 0.9027   & 32.77 / 0.9008 &  32.75 / 0.9027 &  \color{red}{33.15 / 0.9088} &  32.91 / 0.9046 & \color{blue}{32.92 / 0.9048} \\
                          & 25    & 29.97 / 0.8505   & 30.05 / 0.8515   & 30.44 / 0.8618   & 30.38 / 0.8601 & 30.43 / 0.8634 &  \color{red}{30.79 / 0.8711} &  30.58 / 0.8665 & \color{blue}{30.72 / 0.8689} \\
                          & 50    & 26.72 / 0.7676   & 26.82 / 0.7677   & 27.18 / 0.7827   & 27.14 / 0.7804 & 27.32 / 0.7903 & \color{red}{27.74 / 0.8056} & 27.52 / 0.7968 & \color{blue}{27.62 / 0.7999} \\
\multirow{3}{*}{BSD68}    & 15    & 31.08 / 0.8722 & 31.42 / 0.8822 & 31.73 / 0.8906 & 31.63 / 0.8881 &  31.63 / 0.8902 & 31.86 / 0.8947 & \color{blue}{32.00 / 0.8961} & \color{red}{32.02 / 0.8961} \\
                          & 25    & 28.57 / 0.8017 & 28.92 / 0.8148 & 29.23 / 0.8278 & 29.15 / 0.8249 &  29.19 / 0.8289 &  29.41 / 0.8360 & \color{blue}{29.51 / 0.8370} & \color{red}{29.59 / 0.8392} \\
                          & 50    & 25.62 / 0.6869 & 25.97 / 0.7021 & 26.23 / 0.7189 & 26.19 / 0.7171& 26.29 / 0.7245 &   26.53 / 0.7366 & \color{blue}{26.61 / 0.7371} & \color{red}{26.66 / 0.7396} \\
\multirow{3}{*}{Urban100} & 15    & 32.34 / 0.9220 & 31.98 / 0.9187 & 32.67 / 0.9250 & 32.49 / 0.9244 & 32.43 / 0.9273 &   \color{red}{33.17 / 0.9357} &   32.85 / 0.9308 & \color{blue}{32.90 / 0.9315} \\
                          & 25    & 29.70 / 0.8777 & 29.29 / 0.8731 & 29.97 / 0.8792 & 29.82 / 0.8839 & 29.92 / 0.8886 &   \color{red}{30.66 / 0.9026} &  30.34 / 0.8947 &   \color{blue}{30.63 / 0.8999} \\ 
                          & 50    & 25.94 / 0.7791 & 25.71 / 0.7756 & 26.28 / 0.7869 & 26.14 / 0.7927 & 26.52 / 0.8057 &  \color{red}{27.42 / 0.8371} &  27.12 / 0.8239 & \color{blue}{27.34 / 0.8297} \\ \bottomrule
\end{tabular}
\end{table*}

In this subsection, experiments are conducted to compare the proposed model with current CNN-based denoising models with competitive performance. As shown in Fig. \ref{fig_vis_1} and Fig. \ref{fig_vis_2}, the results processed by the proposed model contains more details and clearer edges than DnCNN, IRCNN and FFDNet, and the results from IGNet is similar to that of MWCNN, yet with much smaller model size. The results indicates that the proposed model has the capability to preserve edges and details in denoising procedure due to the guidance of inter-frequency information, which provides promising results in the test images.

\begin{figure} 
\centering 
\includegraphics[width=0.5\textwidth]{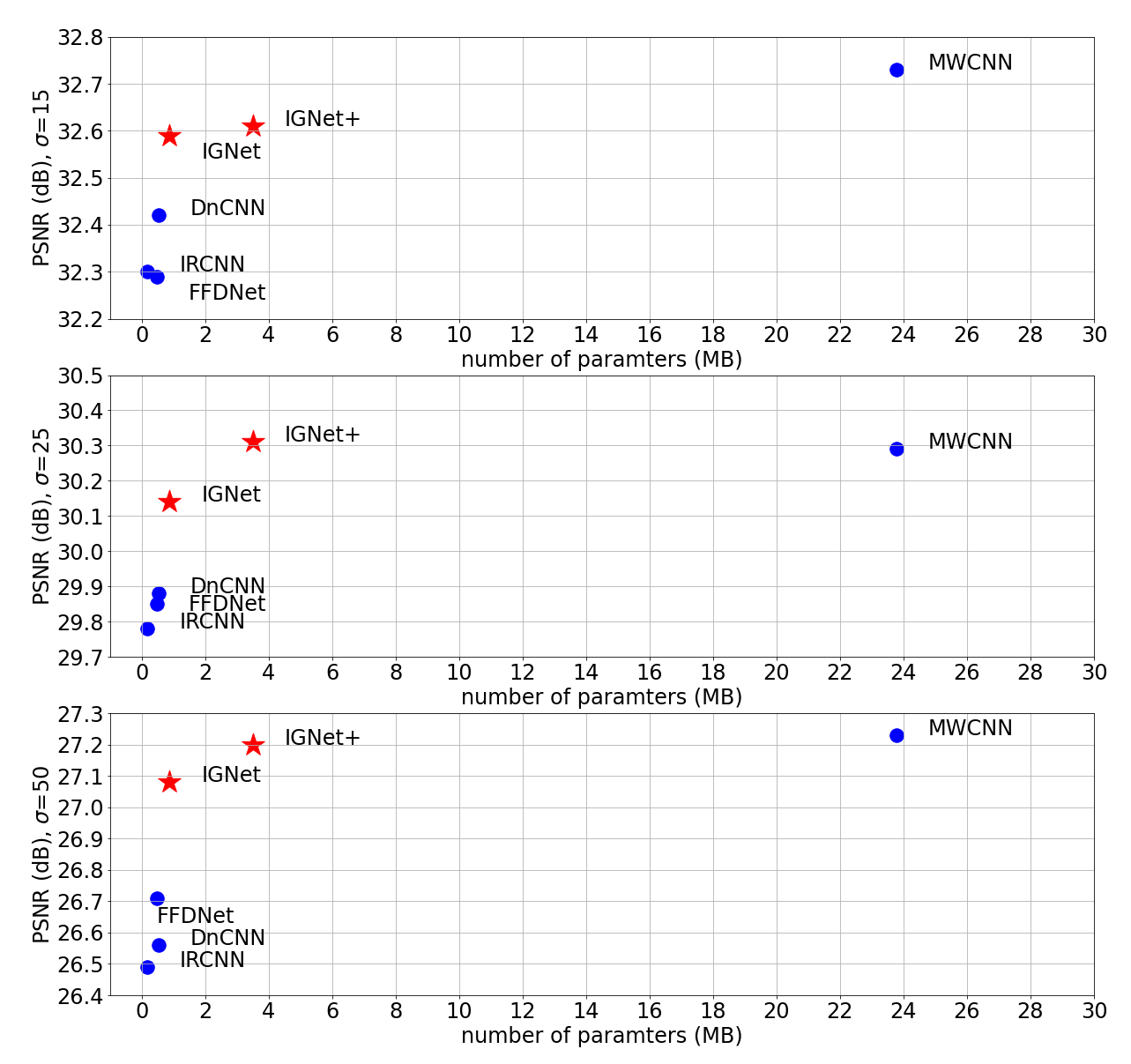} 
\caption{Average PSNR in each noise level vs. model size measured in number of parameters} 
\label{fig_psnr_vs_params} 
\end{figure}

In order to evaluate the performance of proposed method quantitatively, we train the IGNet for image denoising task with different noise levels, and compare the results with previous methods. The methods chosen are four well-performed CNN-based methods: DnCNN \cite{zhang2017beyond}, IRCNN \cite{zhang2017learning}, FFDNet \cite{zhang2018ffdnet}, MWCNN \cite{liu2018multi}. The evaluation metrics used for comparisons are PSNR (Peak Signal-to-Noise Ratio) and SSIM (Structural SIMilarity index) as in previous works. Table \ref{tab_sota_compare} shows the results of these selected comparing methods and proposed method. We also implement the proposed model with more feature maps and the same architecture, which will be denoted as IGNet+ in the following statements. As can be seen from the table, our IGNet+ reached best performance in BSD68 dataset compared with all other methods, and second best performance in Set12 and Urban100 than only behind MWCNN. Moreover, the comparison of number of parameters used to learn denoising task for different models has been shown in Table. \ref{tab_param_compare} and illustrated in Fig. \ref{fig_psnr_vs_params} in a more intuitive manner. The x-axis of Fig. \ref{fig_psnr_vs_params} refers to the number of learned parameters of the model, and the y-axis shows the performance measured in PSNR under the corresponding number of parameters. The PSNRs are averaged in all three test sets mentioned above. Comparing other methods, the proposed IGNet and IGNet+ requires less parameters to obtain a competitive result than other methods. IGNet has similar number of parameters to DnCNN, yet has about 0.17 dB, 0.26 dB and 0.52 dB higher PSNR (in $\sigma=15,25,50$ respectively). Moreover, MWCNN surpasses IGNet+ with about 0.12dB in $\sigma=15$, and has a slight difference of 0.02dB and 0.03dB in noise level settings of $\sigma=25$ and $\sigma=50$, but the number of parameters are about 7 times larger when compared with IGNet+.
The lightweight property of network is considered to be attributed to the exploitation of inter-frequency prior of images. The information inside the image itself helps to reduce the burden of learning them using a large model parameter set. 

\begin{table}[]
\scriptsize
\centering
\caption{Comparisons of average performance vs. number of parameters of each model architecture.}
\label{tab_param_compare}
\begin{tabular}{@{}ccccccc@{}}
\toprule
             & DnCNN & IRCNN & FFDNet  & MWCNN & IGNet & IGNet+ \\ \midrule
\# params (MB) & 0.53  & 0.18  & 0.46   & 23.77 & 0.87  & 3.49 \\
$\sigma=15$ & 32.42 & 32.30 & 32.29  & 32.73 & 32.59  & 32.61 \\
$\sigma=25$ & 29.88 & 29.78 & 29.85   & 30.29 & 30.14 & 30.31 \\
$\sigma=50$ & 26.56 & 26.49 & 26.71  & 27.23 & 27.08 & 27.20 \\ \bottomrule
\end{tabular}
\end{table}

\section{Conclusion}
This paper presents a novel network architecture named IGNet to explore the benefits from using inter-frequency prior of images in as guidance for Image denoising task. The core idea of this network is to split the features into different frequency subbands using DWT, and deal with low-frequency features $LL$ and high-frequency features \{$HL$, $LH$, $HH$\} separately, where low-frequency features continue to be split by DWT, and serve as a high fidelity guidance, while high-frequency features take use of low-frequency guidance for progressive refinement. Attributed to the advantages from internal inter-frequency similarity prior of images, our model can be very lightweight yet get competitive performance. Ablation studies and experiments show both the effectiveness and small size of proposed network.

\ifCLASSOPTIONcaptionsoff
  \newpage
\fi





\bibliographystyle{IEEEtran}
\bibliography{IEEEabrv, Bibliography}

\vfill


\end{document}